\documentclass[letterpaper]{article} 
\usepackage{aaai25}  
\usepackage{times}  
\usepackage{helvet}  
\usepackage{courier}  
\usepackage[hyphens]{url}  
\usepackage{graphicx} 
\usepackage{natbib}  
\usepackage{caption} 
\usepackage{algorithm}
\usepackage{algorithmic}

\usepackage{multirow}
\usepackage{amsmath}
\usepackage{mathrsfs}
\usepackage{mathtools}
\usepackage{booktabs}
\usepackage{tabularx}
\usepackage{xcolor}
\usepackage{array}
\usepackage{colortbl}

\usepackage{newfloat}
\usepackage{listings}
\DeclareCaptionStyle{ruled}{labelfont=normalfont,labelsep=colon,strut=off} 
\floatstyle{ruled}
\newfloat{listing}{tb}{lst}{}
\floatname{listing}{Listing}
\title{Activated Parameter Locating via Causal Intervention for Model Merging}
\author{
    Fanshuang Kong\textsuperscript{\rm 1}, 
    Richong Zhang\textsuperscript{\rm 1,2}\thanks{Corresponding author}, 
    Ziqiao Wang\textsuperscript{\rm 3}
}
\affiliations{
    \textsuperscript{\rm 1}SKLSDE, School of Computer Science and Engineering, Beihang University, Beijing, China\\
    \textsuperscript{\rm 2}Zhongguancun Laboratory, Beijing, China \\
    \textsuperscript{\rm 3}School of Electrical Engineering and Computer Science, University of Ottawa, Ottawa, Canada \\
    kongfs@buaa.edu.cn, zhangrc@act.buaa.edu.cn, zwang286@uottawa.ca
}

\usepackage{bibentry}

\begin{document}

\maketitle

\begin{abstract}
Model merging combines multiple homologous models into one model, achieving convincing generalization without the necessity of additional training. A key challenge in this problem is resolving parameter redundancies and conflicts across multiple models. Existing models have demonstrated that dropping a portion of delta parameters can alleviate conflicts while maintaining performance. However, these methods often drop parameters either randomly or based on magnitude, overlooking task-specific information embedded in fine-tuned models. In this paper, we propose an Activated Parameter Locating (APL) method that utilizes causal intervention to estimate parameter importance, enabling more precise parameter drops and better conflict mitigation. Moreover, to reduce the computational complexity associated with a large number of parameter partitions, we also introduce a theoretically supported gradient approximation strategy for APL. Experiments on model merging within both in-domain and out-of-domain settings, along with associated analyses, showcase the effectiveness of APL.

\end{abstract}

%

\section{Introduction}
Recently, large language models (LLM) have achieved significant breakthroughs across various NLP tasks, with numerous task-specific checkpoints, fine-tuned on open-source LLMs, now publicly available \cite{wolf2019huggingface}. These fine-tuned models incorporate valuable, high-quality task-specific information on top of the pre-trained models \cite{devlin2018bert}. However, obtaining supervised task-specific data remains challenging due to security and privacy concerns, while training LLMs on extensive datasets is also costly \cite{hu2021lora}.
Thus, a method that directly leverages the capabilities of these fine-tuned models with minimal or no additional training would be highly beneficial.
Another relevant concept is linear model connectivity \cite{frankle2020linear}, which suggests that fine-tuned models originating from the same pre-trained model, even when using different hyperparameter settings, generally lie within a single low-error basin \cite{izmailov2018averaging_weights,kuditipudi2019explaining,garipov2018loss}. Merging these models in parameter space can identify points that provide improved generalization \cite{wortsman2022model_model_soup}. Building on this theory, several studies have explored the problem of model merging, where multiple homologous models are combined to achieve better generalization without necessitating a large volume of high-quality data to fully fine-tune LLMs \cite{goddard2024arcee,stoica2023zipit,huang2023lorahub}.

\begin{figure}
    \centering
    \includegraphics[width=\linewidth]{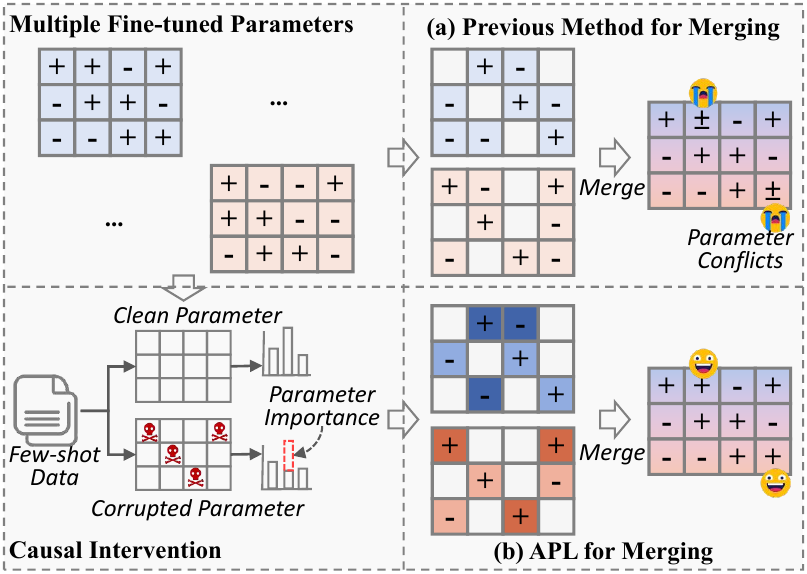}
    \caption{Illustration of APL. `+' and `-' represent essential conflicts in fine-tuned parameters, and `±' represents the conflict while merging.}
    \label{fig:motivation}
\end{figure}

A backbone of model merging involves manipulating and merging the task vector \cite{ilharco2022editing_task_arithmetic}, also known as delta parameters (see Eq.~\ref{eq:delta-para}), which quantify the difference between a pre-trained model and its fine-tuned counterpart. Clearly, delta parameters contain task-specific knowledge learned during fine-tuning, reflected in the shifts within the parameter space. 
Furthermore, while \cite{yu2024language_dare,yadav2024ties} highlight the critical role of delta parameters in model merging, they also observe a major challenge: resolving redundancies and conflicts among the delta parameters of multiple models, particularly as the number of models being merged increases, which tends to aggravate these conflicts. 
Model pruning methods demonstrate that dropping a portion, even a significant portion, of model parameters does not necessarily degrade the performance \cite{frankle2018lottery,srinivas2015data}. 
Leveraging this insight, \cite{yu2024language_dare,yadav2024ties} reduce redundant parameters to mitigate conflicts by zeroing out some delta parameters, based on their magnitudes or by random selection. However, these methods rely solely on statistical information to decide which parameters to eliminate, neglecting the task-specific information embedded in different fine-tuned models.

In this work, we first propose an activated parameter locating (APL) method using causal intervention to estimate parameter importance. We then calibrate the drop ratio and merging weight based on different levels of this importance. Figure \ref{fig:motivation} illustrates the difference between APL and previous models. Notably, APL utilizes few-shot samples for each task, which, as noted by \citet{xiao2023lm_cocktail}, is a moderate condition. These limited samples significantly improve the ability to identify parameters most relevant to a specific task.

Specifically, motivated by \citet{meng2022locating_rome}, which suggests that neuron activations could be identified by the causal intervention on model factual prediction, we use causal tracing to analyze whether the parameter is more important than others based on the few-shot data. In the causal tracing mechanism, we conduct two runs: a clean run, where the fine-tuned model makes predictions as usual, and a corrupted run, where certain parameters of the fine-tuned model are substituted by corresponding pre-trained parameters. The difference in predictions between these two runs reflects the difficulty of the corrupted model in replicating the clean model's performance. This difference reveals the task-specific information embedded in the corrupted parameters, thus indicating their importance for the task. By partitioning parameters at different levels---such as model-level, layer-level, and hidden state-level---we can estimate the importance of each partition and identify the activated parameters. Moreover, to reduce the computational burden of causal tracing when dealing with a large number of parameter partitions, we also introduce a gradient-based approximation method for evaluating parameter importance. This method uses the absolute inner product between the task vector (i.e., delta parameters) and the pre-trained model's gradient vector for specific parameter partitions to approximate their importance. The effectiveness of this speed-up method is theoretically justified.

Furthermore, the parameter importance facilitates model merging from two perspectives: first, by calibrating the drop ratio for different parameter partitions based on their layer-level or hidden state-level importance, and second, by adjusting the merging weight based on model-level importance.
Intuitively, a higher importance for a parameter partition indicates a greater amount of task-specific information contained within it, suggesting that it should retain more neurons during parameter pruning. Conversely, partitions with lower importance can be pruned more aggressively. To fully invoke the relative relationships of parameter importance, we apply a $\tanh$ activation function to map the importance scores, guiding the adjustment of drop ratios accordingly.
This enables APL to prune parameters more extensively while maintaining comparable performance, hence reducing conflicts during model merging.
On top of that, model-level importance reflects the total amount of task-specific information in the fine-tuned model, and also indicates the generalization ability of the pre-trained model. A higher model-level importance suggests that a fine-tuned model should receive a larger merging weight. We adjust the merging weight while preserving the relative scale of model importance based on this principle.

In summary, our contributions are as follows:
\begin{itemize}
    \item We introduce an Activated Parameter Locating (APL) method using causal intervention to estimate parameter importance. This approach helps the precise removal of redundant parameters, effectively mitigating conflicts during model merging.
    \item To mitigate the computational complexity of causal intervention when dealing with a large number of parameter partitions, we develop a gradient approximation variant of APL, with theoretical support for its effectiveness.
    \item Experiments in both in-domain and out-of-domain model merging and related analysis demonstrate the effectiveness of APL.
\end{itemize}

\section{Related Works}

Multiple models fine-tuned with different hyperparameter configurations from one pre-trained model tend to appear to lie in a single low-error basin \cite{izmailov2018averaging_weights}.  Averaging these models leads to better generalization than conventional training \cite{wortsman2022model_model_soup}. This is called linear mode connectivity. Based on this, model merging, which involves merging multiple homologous models, can achieve competitive in-domain task accuracy and reasonably out-of-domain task generalization \cite{yadav2024ties,jin2022dataless_regmean}. Beyond simply averaging multiple model parameters \cite{wortsman2022model_model_soup}, there are many merging methods have been proposed. 
Fisher Merging \cite{matena2022fishermerging} computes the weight importance for each model by the Fisher Information matrix. 
RegMean \cite{jin2022dataless_regmean} uses local linear regression to minimize prediction differences between the merged and individual models, with the closed-form solution providing the merging weight.
Task Arithmetic \cite{ilharco2022editing_task_arithmetic} proves that applying arithmetic operations on delta parameters could steer the merging model. 
Ties \cite{yadav2024ties} retains delta parameters with large magnitude and solves the sign conflict through an electing mechanism.
Dare \cite{yu2024language_dare} reduces potential conflicts in the delta parameters by randomly dropping.
\citet{daheim2023model_uncertainty} links inaccuracies in weighted averaging to gradient discrepancies and proposes an uncertainty-based method to enhance performance by minimizing these discrepancies.
Model Stock \cite{jang2024model_Stock} identifies a connection between performance and proximity to the weight space center, approximating a center-close weight with just two fine-tuned models.

\section{Preliminaries}

\paragraph{Problem Setup}
Let $\mathbf{\Theta}_b$ denote the parameters of a pre-trained base model, and let
$\mathbf{\Theta}_t$ denote the model fine-tuned from $\mathbf{\Theta}_b$ for task $t$.
Given the pre-trained $\mathbf{\Theta}_b$ and a set of $T$ fine-tuned models 
$\{\mathbf{\Theta}_{t_1}, \mathbf{\Theta}_{t_2}, \mathbf{\Theta}_{t_3}, ..., \mathbf{\Theta}_{t_T}\}$,
model merging aims to merge these homologous models to generate a new model,  $\widetilde{\mathbf{\Theta}}$, that can perform well across all $T$ in-domain tasks simultaneously or generalize effectively to an out-of-domain task $t_o$.

\paragraph{Delta Parameter and Parameter Pruning} The delta parameter, also known as the task vector, is defined as the difference between the fine-tuned model and the pre-trained model in parameter space. Formally, given $\mathbf{\Theta}_b$ and $\mathbf{\Theta}_t$, 
the delta parameter of task $t$ is defined as:
\begin{equation}
\label{eq:delta-para}
    \mathbf{\Delta}_t = \mathbf{\Theta}_t - \mathbf{\Theta}_b.
\end{equation}

\citet{yu2024language_dare} and \citet{yadav2024ties} reveal the significant redundancy and sign conflicts in delta parameters across different tasks, suggesting that pruning these parameters is crucial for effective model merging. A naive delta parameter pruning method, proposed by \citet{yu2024language_dare}, involves randomly dropping delta parameters with a global drop ratio $\lambda\in[0,1]$,  followed by rescaling the remaining parameters by a factor of $1/(1 - \lambda)$ to maintain the same expected value of the original embeddings. This process is formulated as:
\begin{equation}
\begin{aligned}
    \mathbf{M}_t & \sim {\rm Bernoulli}(\lambda), \\
    {\mathbf{\Delta}_t} & =  (1 - \mathbf{M}_t) \odot {\mathbf{\Delta}_t}, \\
    {\mathbf{\Delta}_t} & =  {\mathbf{\Delta}_t} / (1 - \lambda),
\end{aligned}
\end{equation}
where $\mathbf{M}_t$ is a mask matrix corresponding to $\mathbf{\Delta}_t$, sampled from a Bernoulli distribution with mean $\lambda$.

\section{Activated Parameter Locating}
As previously discussed, reducing redundancies and conflicts in the delta parameters of multiple fine-tuned models is crucial for effective model merging \cite{yadav2024ties,yu2024language_dare}. 
Current methods typically remove redundant parameters either randomly or based on numerical magnitude, which, although efficient, often overlook essential task-specific information embedded in each fine-tuned model.
In practical scenarios, obtaining few-shot samples for the target task is often feasible \cite{xiao2023lm_cocktail}, and these samples can provide valuable task-specific information through inference or gradient signals. This information can significantly improve the identification of redundant parameters and the minimization of conflicts during model merging. In addition, we note that the associated memory and time costs of using few-shot samples are acceptable, considering the potential benefits of redundancy reduction. In this case, the key challenge is to effectively identify redundant delta parameters using few-shot samples, or more specifically to locate task-specific activated parameters in a fine-tuned model.

Motivated by \citet{meng2022locating_rome}, which uses causal tracing to identify neuron activations crucial for a model's factual predictions, we develop a causal intervention method to locate activated parameters using few-shot samples from a specific task. In particular, given the few-shot sample $D_t$ of target task $t$, we first estimate the parameter importance of different parameter partitions by causal tracing factual associations. We then use different levels of importance to calibrate the drop ratio for each parameter partition and adjust the merging weight. By incorporating task-specific information from few-shot samples, our Activated Parameter Locating (APL) method can more accurately identify redundant parameters, thereby alleviating parameter conflicts and improving the effectiveness of model merging.

\subsection{Causal Intervention on Activations}

\paragraph{Information Flow of Causal Tracing} 
Analyzing the information flow in causal tracing can help determine the relative importance of specific states, as suggested by \citet{meng2022locating_rome}. To quantify the contribution of parameters to few-shot samples in a specific task, we conduct two activation analysis runs: 
  a clean run using the fine-tuned model $\mathbf{\Theta}_t$ to predict the true label probability for $D_t$, and a corrupted run where $\mathbf{\Theta}_t$ is intervened with ``noise'' parameters to evaluate the ability of parameters for information restoration. 

Specifically, in the clean run, the few shot sample $D_t$ is fed into the corresponding fine-tuned model $\mathbf{\Theta}_t$, delivering the true label probability $P_t$. We denote the information flow as $\rightarrow$, and the process is represented as follows:
\begin{equation}
    D_t \rightarrow \mathbf{\Theta}_t \rightarrow P_t.
\end{equation}

In the corrupted run, 
the fine-tuned model is modified by replacing certain subset of parameters, denoted as $\mathbf{\Theta}_t^{[*]}$, with their corresponding ``noise'' parameters from the pre-trained model, denoted as $\mathbf{\Theta}_b^{[*]}$. The sample $D_t$ is then passed through this corrupted model, producing a prediction output denoted as $P_t^*$. The information flow for the corrupted run is represented as:
\begin{equation}
    D_t \rightarrow \mathbf{\Theta}_t^{[: \setminus *]} \oplus \mathbf{\Theta}_b^{[*]} \rightarrow P_t^*,
\end{equation}
where $\oplus$ denotes concatenation operation, and $\mathbf{\Theta}_t^{[: \setminus *]}$ represents the fine-tuned parameters $\mathbf{\Theta}_t$ excluding $\mathbf{\Theta}_t^{[*]}$.

The clean run evaluates the impact of all delta parameters on a specific task, while the corrupted run evaluates only a subset of these parameters. The difference 
between $P_t$ and $P_t^*$ quantifies the difficulty of recovering from the corrupted run to the clean run. By assessing this difficulty, we can estimate the importance of corrupted delta parameters for a specific task.

\paragraph{Parameter Importance}
Formally, the parameter importance, denoted as $I_t^*$, 
is defined as the difference between the perturbed prediction $P_t^*$ and the original prediction $P_t$:
\begin{equation}
    I_t^* = P_t^* - P_t.
\end{equation}
Note that $P_t^*$ is typically smaller than $P_t$, as replacing components in well-trained $\mathbf{\Theta}_t$ generally does not improve the prediction performance. In this case, a smaller value of $I_t^*$ (i.e. a larger gap between $P_t^*$ and $P_t$) indicates greater difficulty in reconstructing accurate predictions, implying that the corrupted parameters have a stronger causal impact on the prediction. 
This suggests that these corrupted parameters are crucial for the task $t$ and should not be removed while pruning. Consequently, tracing causality to determine parameter importance allows us to identify the activated parameters for a specific task.

\subsection{Coarse Grained Parameter Partition}
Ideally, causal intervention would allow us to obtain the importance of each neural activation
or parameter. However, in practical applications, evaluating parameter-level importance is computationally expensive. To address this, we partition parameters into model-level, layer-level and hidden state-level based on the network structure. For model-level parameter importance, the entire fine-tuned model is intervened by the pre-trained model, as formulated in Eq.~\ref{eq:model-level}. The importance of the resulting model, denoted as $I_t^m$, represents the complete task-specific information embedded in the fine-tuned model. Later on we will show that this measure is particularly useful for merging multiple models and determining the associated weights for each model, as described in Eq. \ref{eq:in-domain model merging}.
Moreover, for layer-level and hidden state-level parameter importance, denoted as $I_t^l$ and  $I_t^h$, respectively, 
we partition parameters based on layers or hidden states within the model structure. Their corrupted runs are defined by Eq.~\ref{eq:layer-level} and Eq.~\ref{eq:hidden-level}, respectively.

\begin{equation}
    D_t \rightarrow \mathbf{\Theta}_b \rightarrow P_b,
    \label{eq:model-level}
\end{equation}
\begin{equation}
    D_t \rightarrow \mathbf{\Theta}_t^{[1:l]} \oplus \mathbf{\Theta}_b^{[l]} \oplus \mathbf{\Theta}_t^{[l+1:L]} \rightarrow P_t^l,
    \label{eq:layer-level}
\end{equation}
\begin{equation}
    D_t \rightarrow \mathbf{\Theta}_t^{[1:h]} \oplus \mathbf{\Theta}_b^{[h]} \oplus \mathbf{\Theta}_t^{[h+1:H]} \rightarrow P_t^h,
    \label{eq:hidden-level}
\end{equation}
where $l$ represents the $l_{th}$ layer of total $L$ layers in $\mathbf{\Theta}$, and $h$ represents the $h_{th}$ hidden state.

\begin{figure}
    \centering
    \includegraphics[width=\linewidth]{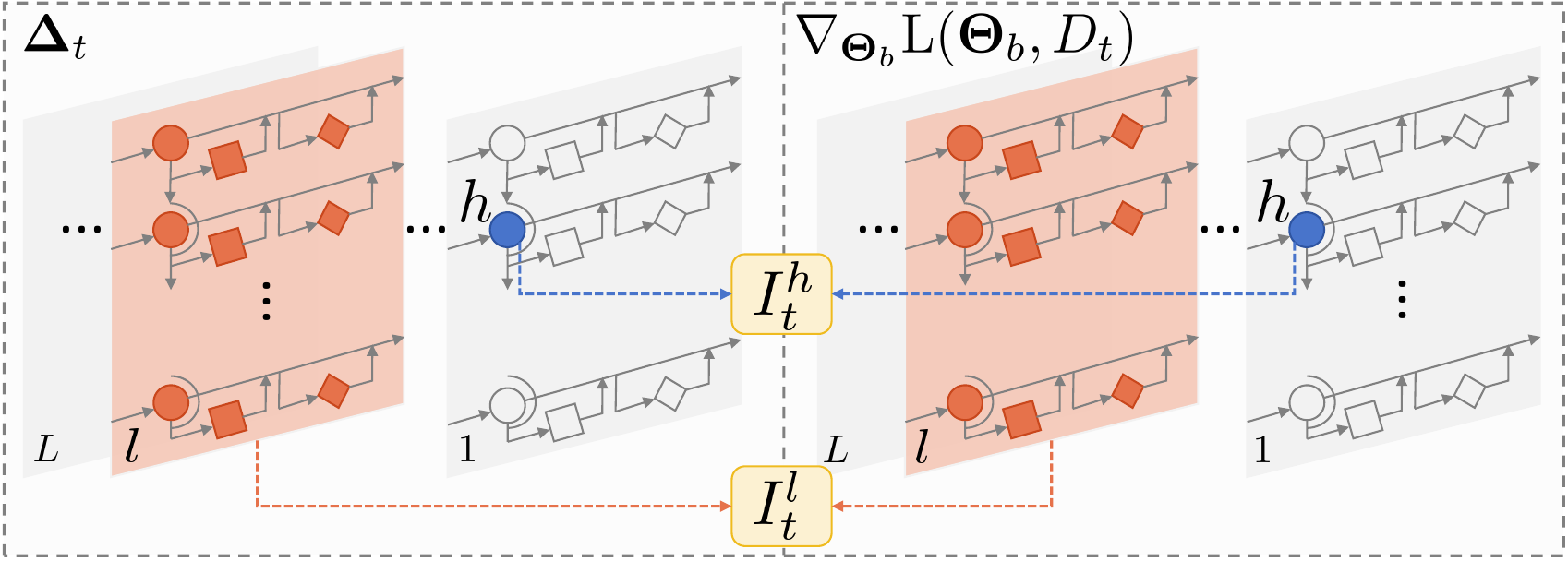}
    \caption{Gradient approximation for APL. The orange and blue components represent the parameters activated at the layer-level and hidden state-level, respectively.}
    \label{fig:model}
\end{figure}

With this out of the way, one can iterate through $l\in[1,L]$ (or $h\in[1:H]$) to compute $I_t^l$ (or $I_t^h$ resp.), ranking the parameter partitions according to their importance. However, when $L$ or $H$ is large, iteratively replacing each parameter partition, even at the layer-level or hidden state-level,  can significantly increase computational complexity. To mitigate this, we also apply a gradient approximation method for causal tracing, particularly when $L$ or $H$ is large. We now explore this.

\subsection{Gradient Approximation for Parameter Importance}

Let $\mathrm{L}(\mathbf{\Theta},D)$ denote the validation loss of a model  $\mathbf{\Theta}$ computed using the few-shot sample $D$. When fine-tuning the pre-trained model $\mathbf{\Theta}_b$ on the few-shot sample $D_t$ for task $t$, the gradient $\nabla_{\mathbf{\Theta}_b}\mathrm{L}(\mathbf{\Theta}_b,D_t)$ characterizes the direction that minimizes the loss $\mathrm{L}(\mathbf{\Theta}_b,D_t)$.
Intuitively, if the partitioned delta parameters in $\mathbf{\Delta}_t$ are orthogonal to the corresponding partition in $\nabla_{\mathbf{\Theta}_b}\mathrm{L}(\mathbf{\Theta}_b,D_t)$, then that partition does not capture task-specific information and can be mostly dropped. 
Consequently, we propose using the magnitude of the inner product between $\mathbf{\Delta}_t$ and $\nabla_{\mathbf{\Theta}_b}\mathrm{L}(\mathbf{\Theta}_b,D_t)$ for each parameter partition to represent its importance. For example, at the layer level $l$, let $\mathbf{\Delta}^l_t$ and $\nabla_{\mathbf{\Theta}_b}\mathrm{L}^l(\mathbf{\Theta}_b,D_t)$ be the delta parameters and the gradient of layer $l$, respectively. The importance of partition $l$ is then given by $I_t^l=|\mathbf{\Delta}^l_t\cdot \nabla_{\mathbf{\Theta}_b}\mathrm{L}^l(\mathbf{\Theta}_b,D_t)|$. In fact, the effectiveness of this gradient-based importance approximation can be theoretically justified, as shown below.

\paragraph{Theoretical Analysis} We apply a first-order Taylor expansion to the loss of $\mathbf{\Theta}_t$ at $\mathbf{\Theta}_b$, and to its pruned version $\mathbf{\Theta}'_t=(1 - \mathbf{M}_t) \odot \mathbf{\Delta}_t + \mathbf{\Theta}_b$, also at $\mathbf{\Theta}_b$:
\begin{align*}
    \mathrm{L}(\mathbf{\Theta}_t, D_t)& \! = \! \mathrm{L}(\mathbf{\Theta}_b, D_t) \! + \! \mathbf{\Delta}_t\cdot \nabla_{\mathbf{\Theta}_b} \mathrm{L}(\mathbf{\Theta}_b, D_t),\\
    \mathrm{L}(\mathbf{\Theta}'_t, D_t)& \! = \! \mathrm{L}(\mathbf{\Theta}_b, D_t) \! + \! (1 \! - \! \mathbf{M}_t) \odot {\mathbf{\Delta}_t} \cdot \nabla_{\mathbf{\Theta}_b} \mathrm{L}(\mathbf{\Theta}_b, D_t).
\end{align*}

Our goal is to find a mask $\mathbf{M}_t$ such that $\mathbf{\Theta}'_t$ performs similarly to the original model $\mathbf{\Theta}_t$. In particular, we aim for $\mathrm{L}(\mathbf{\Theta}'_t, D_t)$ to be close to $\mathrm{L}(\mathbf{\Theta}_t, D_t)$. By equations above,
we derive the difference in loss as follows:
\begin{align}
\label{eq:taylor-difference}
    |\mathrm{L}(\mathbf{\Theta}_t,D_t)\!-\!\mathrm{L}(\mathbf{\Theta}'_t,D_t)|\!\!=\!\! |\mathbf{M}_t \!\odot\! {\mathbf{\Delta}_t} \!\cdot\! \nabla_{\mathbf{\Theta}_b}\! \mathrm{L}(\mathbf{\Theta}_b,D_t)|.
\end{align}

According to Eq. \ref{eq:taylor-difference}, we first note that when $\mathbf{M}_t$ is an all-zero vector, $|\mathrm{L}(\mathbf{\Theta}_t,D_t) - \mathrm{L}(\mathbf{\Theta}'_t,D_t)| = 0$ is trivially satisfied. However, the trivial solution $\mathbf{M}_t = \mathbf{0}$ implies that no parameters are pruned, which is contrary to our objective of dropping as many parameters as possible to reduce redundancy. In fact, RHS of Eq. \ref{eq:taylor-difference} can be reformulated as: 
\begin{align}
    \left|\sum_{i=1}^d\mathbf{M}_t[i] \cdot {\mathbf{\Delta}_t}[i] \cdot \nabla_{\mathbf{\Theta}_b} \mathrm{L}(\mathbf{\Theta}_b,D_t)[i]\right|,
\end{align}
where $d$ is the total number of parameters.
This suggests that if {$\left|\mathbf{\Delta}_t[i] \cdot \nabla_{\mathbf{\Theta}_b} \mathrm{L}(\mathbf{\Theta}_b,D_t)[i]\right|$} is small (i.e., close to zero) for a given dimension $i$, setting $\mathbf{M}_t[i] = 1$ will not increase $|\mathrm{L}(\mathbf{\Theta}_t,D_t) - \mathrm{L}(\mathbf{\Theta}'_t,D_t)|$ significantly, allowing us to remove the delta parameter in this dimension. Conversely, if  $\left|{\mathbf{\Delta}_t}[i]\cdot\nabla_{\mathbf{\Theta}_b} \mathrm{L}(\mathbf{\Theta}_b, D_t)[i]\right|$ is large, we set $\mathbf{M}_t[i] = 0$, indicating that the parameter in this dimension should be retained. Extending this argument to the layer or hidden state level (i.e. parameter partition) is straightforward. Consequently, we use the inner product between the delta parameter and gradient vector of the pre-trained model to quantify the parameter importance of each component in the model, namely $I_t^*=|\mathbf{\Delta}^*_t \cdot \nabla_{\mathbf{\Theta}_b} \mathrm{L}^*(\mathbf{\Theta}_b, D_t)|$. 

Note that incorporating higher-order terms in the Taylor expansion, such as the Hessian matrix, would obtain a more accurate approximation. However, the computational complexity involved in calculating these terms is substantial. Therefore, we only consider a first-order approximation in our approach, which only requires a single backpropagation step for the pre-trained model, and is faster than iteratively replacing parameter partitions when dealing with a large number of partitions.  Figure~\ref{fig:model} further illustrates this gradient-based importance approximation approach.

\subsection{Parameter Importance Guided Drop Ratio}
As stated, 
partitions with higher $I_t^*$ 
contain more task-specific information, suggesting the need to retain more delta parameters and keep the corresponding drop ratio $\lambda_t^*$ relatively low.
To maintain the relative importance relationships and further leverage the absolute importance value for each partition, we introduce a simple $\tanh$ activation function to calibrate the primary drop ratio $\lambda$ based on $I_t^*$. The calibrated drop ratio $\lambda_t^*$ for each parameter partition is defined as follows:
\begin{equation}
\begin{aligned}
    \beta & =  \tanh (I_t^* / \tau_1), \\
    \lambda_t^* & = \begin{cases}
                \lambda - \epsilon, & \text{if } \beta < -\epsilon, \\
                \lambda + \beta, & \text{if } -\epsilon \leq \beta \leq \epsilon, \\
                \lambda + \epsilon, & \text{if } \beta > \epsilon,
                \end{cases}
\end{aligned}
\end{equation}
where $\tau_1$ is the temperature of $\tanh$, and $\epsilon$ is a threshold that determines the upper and lower limits of the adjustment from $\lambda$ to $\lambda_t^*$.

\subsection{APL for In-domain Model Merging}
APL reduces redundant delta parameters more accurately by parameter importance guided drop ratio and therefore mitigates parameter conflicts more efficiently while merging. Additionally, model-level importance measures the entire task-specific information learned by the fine-tuned model compared to the pre-trained model. A smaller model importance indicates that the pre-trained model 
already generalizes well to the task, whereas a larger value suggests that the task contains more specific knowledge for the pre-trained model to learn. When merging multiple models, the fine-tuned model with higher importance should have a higher proportion to retain more task-specific information.

Accordingly, we first apply APL 
to prune the delta parameters for the fine-tuned models, and then utilize the model importance directed weight to merge multiple models, represented as:
\begin{equation}
    \begin{aligned}
        w_t^m & = {\rm softmax}(I_t^m / \tau_2), \\
        \mathbf{M}_t^* & \sim {\rm Bernoulli}(\lambda_t^*),  \\
        \widetilde{\mathbf{\Theta}} \; & = \sum_{t=1}^{T}{w_t^m(\mathbf{\Theta}_b + (1 - \mathbf{M}_t^*) \odot \mathbf{\Delta}_t)},
    \end{aligned}
    \label{eq:in-domain model merging}
\end{equation}
where $\tau_2$ is the temperature of the softmax activation function, and $w_t^m$ is the weight of target task $t$ to merge. In practice, a straightforward approach is to directly merge pruned APL models, which is equivalent to applying APL on the Task Arithmetic \cite{ilharco2022editing_task_arithmetic} method. Another improved approach is to introduce enhanced model importance guided merging weight, which we refer to as MI-Task Arithmetic in experiments.

\subsection{APL for Out-of-domain Model Merging}
In the out-of-domain model merging setting, the input of causal tracing information flow is the few-shot sample $D_o$ of the unseen task $t_o$. The problem is transformed into selecting relevant delta parameters from each fine-tuned model based on $D_o$ to obtain the ability to solve the task $t_o$.
Other than the changes in few-shot data, the rest of the process remains consistent with the in-domain setting, referring to Eq. \ref{eq:in-domain model merging}.

\section{Experiments}
\subsection{Experiments Setup and Implementation Details}
\noindent \textbf{Datasets and Backbone of Merging Model} Following \citet{xiao2023lm_cocktail}, we adopt Llama-2-chat-7B \cite{touvron2023llama} as the pre-trained model. And we select 6 representative tasks, including AG News, Hellaswag, MNLI, MRPC, SST2 and Winogrande, which utilize Accuracy as evaluation metrics from FLAN \cite{wei2021finetuned_flan} to fine-tune the Llama. All relevant datasets and fine-tuned models could be downloaded from Github\footnote{\url{https://github.com/FlagOpen/FlagEmbedding/tree/master/LM_Cocktail}}. Dataset statistics and description refer to the Appendix.

\noindent \textbf{Evaluation} For the in-domain setting, we merge all separately fine-tuned models into a single one, and evaluate the merging model on all datasets. For the out-of-domain setting, the merging model is evaluated on 57 tasks from MMLU \cite{hendrycks2020measuring_mmlu}, containing STEM, Social Sciences, Humanities and Others. The few-shot samples of both in-domain and out-of-domain tasks are consistent with \cite{xiao2023lm_cocktail}.

\noindent \textbf{Implementation Details} For hyperparameters, we select $\lambda$ from $\{0.5, 0.6, 0.7, 0.8, 0.9, 0.95, 0.99, 0.995\}$, and $\epsilon$ from $\{0.0005, 0.001, 0.005, 0.01, 0.05, 0.1\}$. We set the values of $\tau_1$ and $\tau_2$ to be 5 by a simple parameter grid search. Detailed experimental results are listed in the Appendix. All experiments are conducted on NVIDIA Tesla A100 GPUs. And it takes 900s for APL in the 5-shot setting.

\subsection{Comparable Baselines}
To evaluate the parameter pruning effectiveness, we compare APL with \textbf{Magnitude} \cite{yadav2024ties}  and \textbf{Dare} \cite{yu2024language_dare}. Magnitude drops delta parameters with the smallest absolute change. Dare randomly drops a proportion of delta parameters.
In the in-domain setting, we adopt two popular delta parameter based models for evaluation, including \textbf{Task Arithmetic} \cite{ilharco2022editing_task_arithmetic} and \textbf{Ties} \cite{yadav2024ties}. The Task Arithmetic model naively sums the delta parameters of all tasks with a scale hyperparameter. The SOTA Ties model first prunes the delta parameters using the magnitude method and then mitigates conflicts by an elect mechanism before merging.
In the out-of-domain setting, we mainly compare APL with \textbf{LM-Cocktail} \cite{xiao2023lm_cocktail}, which computes loss via few-shot samples to assign the weight for each fine-tuned model while merging.
\begin{figure}[h]
    \centering
    \includegraphics[width=\columnwidth]{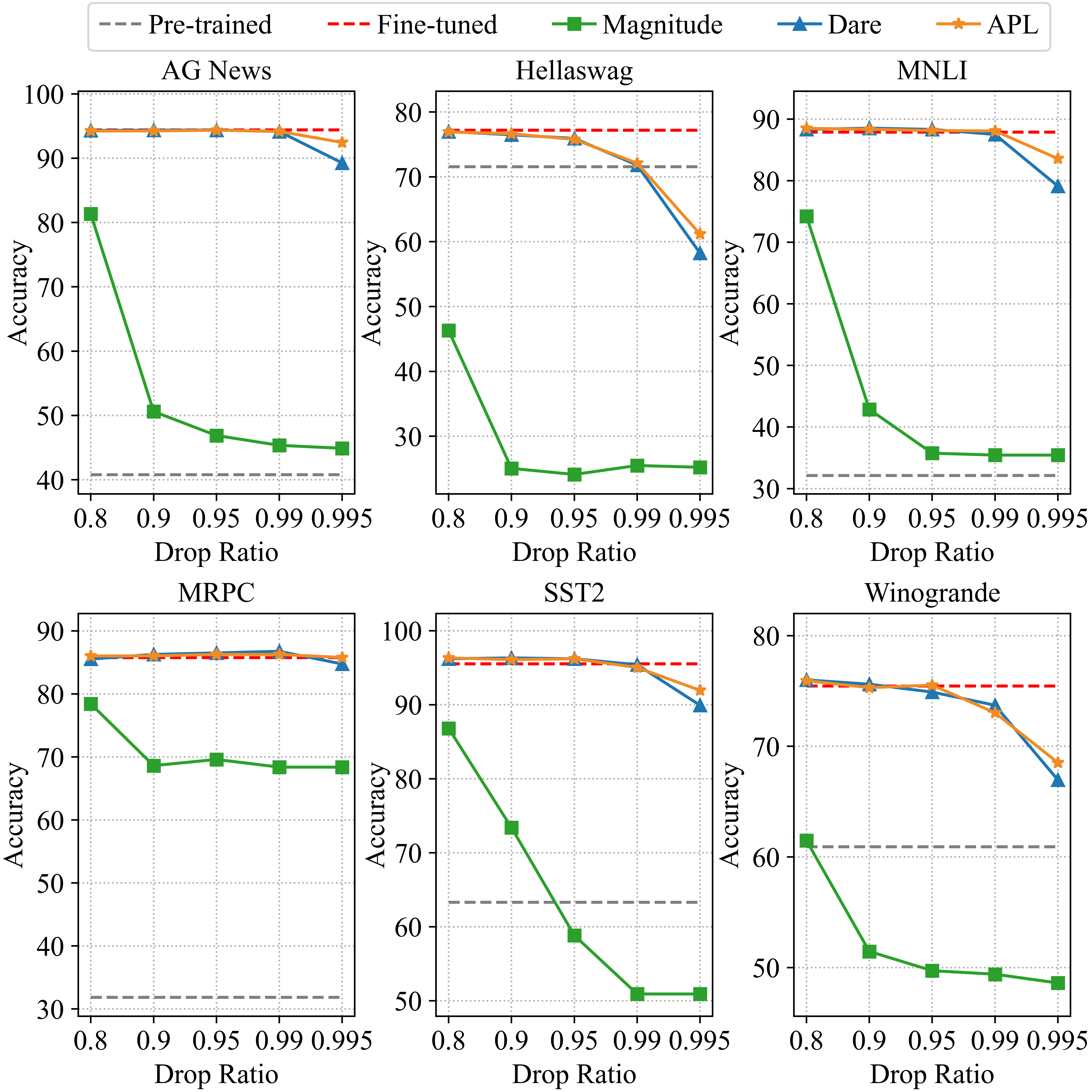}
    \caption{Pruning methods comparison via drop ratio.}
    \label{fig:pruning comparison}
\end{figure}

\begin{table*}
    \centering
    \begin{tabular}{c|cccccc|c}
    \toprule
    Model & AG News & Hellaswag & MNLI & MRPC & SST2 & Winogrande & Average \\
    \midrule
    Pre-trained & 40.80  & 71.58 & 32.14 & 31.86 & 63.30 & 60.93 & 50.10 \\
    Fine-tuned & 94.42  & 77.20 & 87.90 & 85.78 & 95.53 & 75.45 & 86.05 \\
    \midrule
    Ties (SOTA) & 92.43 & 74.98 & 65.42 & 52.20 & 95.06 & 73.16 & 75.54 \\
    \midrule
    Task Arithmetic & \textbf{91.61} & 73.80 & 57.18 & 31.61 & 95.98 & 70.71 & 70.15 \\
    \textit{with Dare} & 90.80 & 75.56 & 57.25 & 31.61 & \textbf{96.90} & 70.40 & 70.42 \\
    \textit{with APL} & 90.97 & \textbf{75.66} & \textbf{57.92} & \textbf{31.62} & 96.79 & \textbf{70.80} & \textbf{70.63} \\
    \midrule
    MI-Task Arithmetic & 88.15 & \textbf{72.06} & 66.06 & 66.91 & \textbf{96.21} & 64.00 & 75.57 \\
    \textit{with Dare} & 88.07 & 72.01 & \textbf{66.11} & 69.36 & 94.72 & 64.01 & 75.71 \\
    \textit{with APL} & \textbf{89.22} & 71.98 & 66.10 & \textbf{70.10} & 95.99 & \textbf{64.09} & \textbf{76.25} \\
    \bottomrule
    \end{tabular}
    \caption{In-domain model merging results. The best performance among all comparable variants is highlighted in bold.}
    \label{tab:in-domain results}
\end{table*}

\subsection{Pruning Comparison: Magnitude vs. Dare vs. APL}
As APL focuses on effectively dropping delta parameters, we compare it with two typical pruning methods, Magnitude and Dare. Figure \ref{fig:pruning comparison} shows the results with different drop ratios. It is worth noting that, to ensure the number of parameters dropped by APL is consistent with Magnitude and Dare, we employ layer-level APL with a linear mapping for parameter importance $I_t^l$ and drop ratio $\lambda_t^l$.
Specifically, when the drop ratios of Magnitude and Dare are equal to $\lambda$, we initially set the drop ratio as $\lambda$ for the layer with the intermediate importance from $I_t^1$ to $I_t^L$. For layers with importance lower than the intermediate value, the drop ratio linearly increases to $\lambda + \epsilon$, while for layers with importance higher than the intermediate value, the drop ratio linearly decreases to $\lambda - \epsilon$. While this linear relationship loses the information on the absolute differences in parameter importance, the fact that the total parameter amount for each layer is equal ensures that the number of parameters retained by APL strictly equals the sum of Magnitude and Dare, thereby ensuring a fair comparison.

From Figure \ref{fig:pruning comparison}, we find Dare and APL are consistently superior to Magnitude, which demonstrates that the magnitude of numerical changes is not positively correlated with task-specific information. Additionally, consistent with the conclusion in \cite{yu2024language_dare}, Dare maintains effectiveness without a decrease in performance when the drop ratio is less than 95\%. However, when it exceeds 95\%, especially reaching 99.5\%, the APL's performance is superior to Dare.
This demonstrates that APL indeed locates more activated parameters than random dropping, thereby enhancing the redundancy identification and pruning effectiveness.

Besides the experiments on the large language model Llama, we initially conducted experiments on the Roberta model \cite{liu2019roberta}. We selected the standard GLUE \cite{GLUE} dataset for the experiments, and the results with consistent conclusions are shown in the Appendix.

\subsection{In-domain Model Merging Results}
In the in-domain merging setting, we mainly implement Task Arithmetic (TA) and MI-Task Arithmetic (MI-TA), for each with \textit{Dare} and \textit{APL} variants applying. Except for the basic pre-trained model and fully supervised fine-tuned model on each task, the SOTA model Ties is taken into account. Comparable results are shown in Table \ref{tab:in-domain results}. From the results, we observe that MI-TA outperforms TA significantly, indicating the effectiveness of the model importance guided merging weight adjustment. Additionally, APL performs better than Dare, demonstrating that APL effectively alleviates parameter conflicts compared with Dare. We also conducted experiments on merging less number of models, with results reported in the Appendix.

\begin{table*}
    \centering
    \begin{tabular}{c|ccc|ccc}
    \toprule
    Dataset & Llama & Llama-ICL & Multi-task-learning & LM-Cocktail & \textit{with Dare} & \textit{with APL} \\
    \midrule
    STEM & 34.46 & 37.08 & 29.73 & 37.71 & 38.11 & \textbf{38.20}  \\
    Social Sciences & 53.28 & 52.33 & 33.84 & 55.77 & 56.90 & \textbf{57.11}  \\
    Humanities & 50.83 & 51.71 & 32.91 & 51.61 & 51.69 & \textbf{52.17} \\
    Others & 49.58 & 49.37 & 36.08 & 52.06 & 52.04 & \textbf{52.49}  \\
    \midrule   
    Average & 45.87 & 46.65 & 32.88 & 48.21 & 48.58 & \textbf{48.88} \\
    \bottomrule
    \end{tabular}
    \caption{Out-of-domain model merging results. Comparable results are token from \citet{xiao2023lm_cocktail}. Bolded results indicate the best performance for each group of dateset.}
    \label{tab:out-of-domain results}
\end{table*}

\subsection{Out-of-domain Model Merging Results}
To evaluate the generalization of APL in out-of-domain tasks, except for three basic models, pre-trained Llama, Llama with in-context-learning (ICL) \cite{gao-etal-2021-making}, and multi-task-learning, we consider the most recently LM-Cocktail as baselines. As LM-Cocktail already calibrates weights by few-shot samples, we only apply \textit{Dare} and \textit{APL} on LM-Cocktail for comparison. Results are shown in Table \ref{tab:out-of-domain results}. Both \textit{Dare} and \textit{APL} perform better than LM-Cocktail, demonstrating that parameter pruning can reduce conflicts and enhance merging effectiveness. The superiority of \textit{APL} over \textit{Dare} indicates that \textit{APL} can better locate activated parameters, leading to superior pruning outcomes.

\subsection{Impact of Different Parameter Partition}
\begin{figure}
    \centering
    \includegraphics[width=\columnwidth]{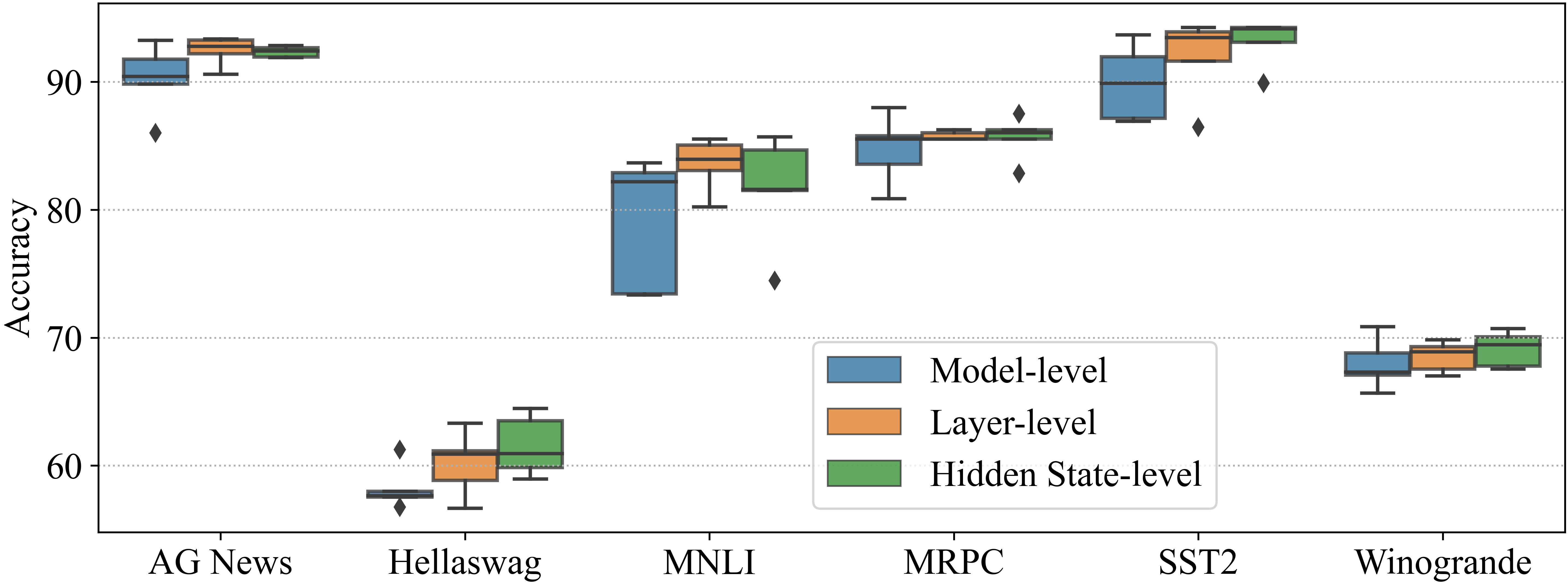}
    \caption{Performance comparison on different levels of parameter partition.}
    \label{fig:different level}
\end{figure}

To evaluate the impact of parameter importance at different levels on activated parameter locating, we compared the pruning effects of APL at the model-level, layer-level and hidden state-level. Noted, the model-level APL adopts the same drop ratio for the overall model, which is equivalent to Dare. Figure \ref{fig:different level} shows the box plots of five runs of experimental results on different datasets. From the Figure, we find that layer-level and hidden state-level APL generally outperforms model-level APL (Dare), providing evidence of the effectiveness of activation locating. The results of hidden state-level APL are better than or approximately equal to those of layer-level APL, indicating that finer-grained parameter partition can locate more effective activated parameters. Moreover, to assess whether gradient approximation significantly affects performance, we conduct experiments detailed in the Appendix. The results indicate that while this approximation has a slight impact on the outcomes, it still delivers considerable performance.
\begin{figure}
    \centering
    \includegraphics[width=\linewidth]{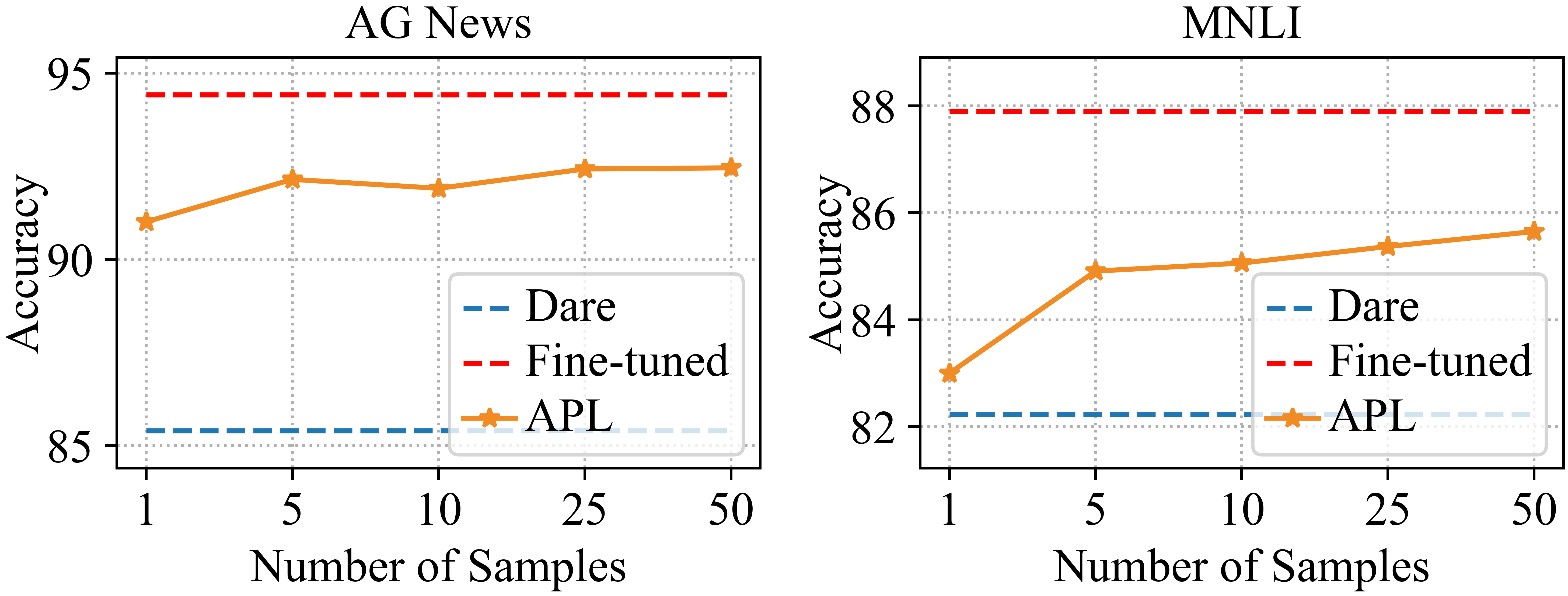}
    \caption{APL performance on AG News and MNLI via different numbers for few-shot samples.}
    \label{fig:few-shot}
\end{figure}
\subsection{Impact of the Number of Few Shot Samples}
To assess whether few-shot samples provide sufficient task-specific information, we present the results of APL with varying numbers of samples in Figure \ref{fig:few-shot}. For different numbers of samples, the upper limit should be the fine-tuned model, which utilizes the whole dataset, and the lower limit should be the Dare model, which corresponds to zero samples. From the Figure, we find more samples achieve higher performance, indicating more task-specific information captured. This is consistent with intuition. Further, just five examples can yield impressive results, which ensures the efficiency of inference and gradient computation. 

\subsection{Activated Parameters Analysis}
\begin{figure}
    \centering
    \includegraphics[width=\linewidth]{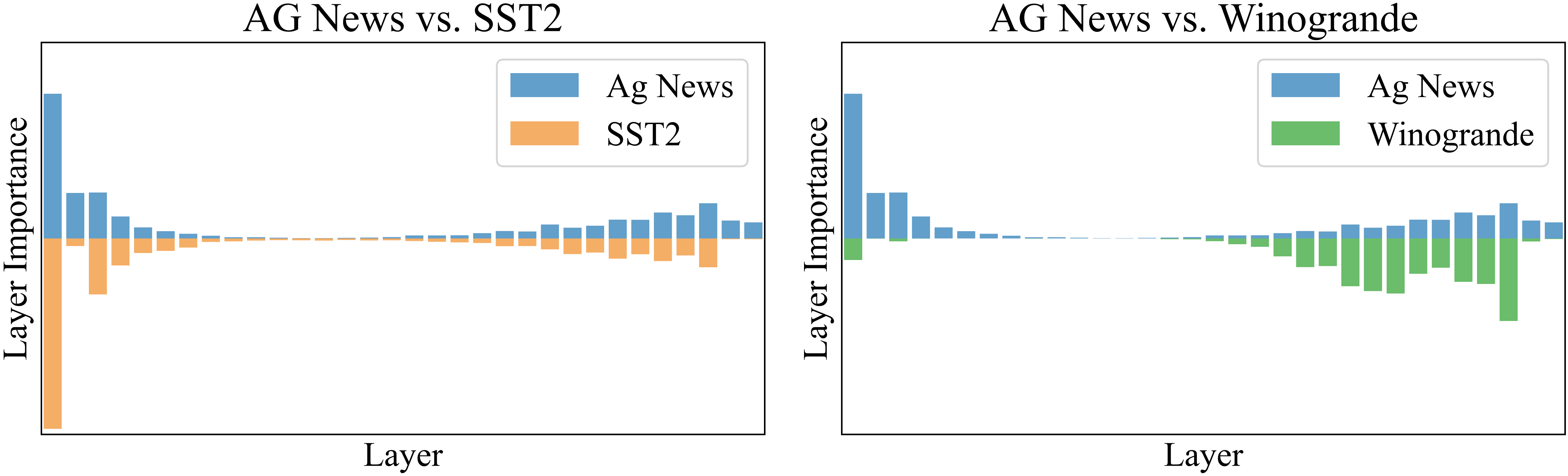}
    \caption{Layer-level parameter importance comparison. Noted, values of the y-axis illustrate differences in distribution rather than indicating positive or negative importance.}
    \label{fig:activation}
\end{figure}

To investigate whether APL locates activated parameters by causal intervention, we derive an analysis of layer-level parameter importance. Specifically, we select two representative classification tasks, AG News and SST2, and a coreference task, Winogrande, to visualize the layer importance via different layers. Results are shown in Figure \ref{fig:activation}. We observe that similar tasks, such as AG News and SST2, tend to activate similar layers, whereas distinct tasks, like AG News and Winogrande, tend to activate different layers. 
This behavior aligns with the principles of language model interpretability \cite{jawahar2019does}, that is, representations of different layers in the language model serve different tasks. This demonstrates APL indeed locates activated parameters for specific tasks.

\section{Conclusion}
Parameter redundancies and conflicts among multiple fine-tuned models are crucial challenges in model merging.
In this paper, we develop an activated parameter locating method to estimate the parameter importance at different levels, including model-level, layer-level, and hidden state-level. Leveraging this importance for drop ratio and merging weight calibration could retain fewer neurons while maintaining comparable performance, further alleviating redundancies and conflicts. Additionally, we theoretically deliver a gradient-based approximation for APL, which alleviates the complexity of causal tracing computation. Experiments and related analyses demonstrate the effectiveness of APL.

\bibliography{aaai25}

\appendix
\newpage
\section{Appendices}
\subsection{Datasets}
\label{app:dataset}
We select six representative datasets from FLAN. Dataset statics are shown in Table \ref{tab:dataset}.

\begin{table}[h]
    \centering
    \resizebox{0.8\linewidth}{!}{
    \begin{tabular}{c|ccc}
        \toprule
        Datasets & \#Train & \#Test & Task \\
        \midrule
        AG News & 30,000 & 7,600 & Summarize \\
        Hellaswag & 30,000 & 10,042 & Commonsense \\
        MNLI & 30,000 & 9,815 & NLI \\
        MRPC & 3,668 & 408 & Paraphrase \\
        SST2 & 30,000 & 872 & Sentiment \\
        Winogrande & 30,000 & 1,267 & Coreference \\
        \bottomrule
    \end{tabular}
    }
    \caption{Dataset statics.}
    \label{tab:dataset}
\end{table}

\begin{figure}[h]
    \centering
    \begin{tabular}{cc}
    \includegraphics[width=0.47\columnwidth]{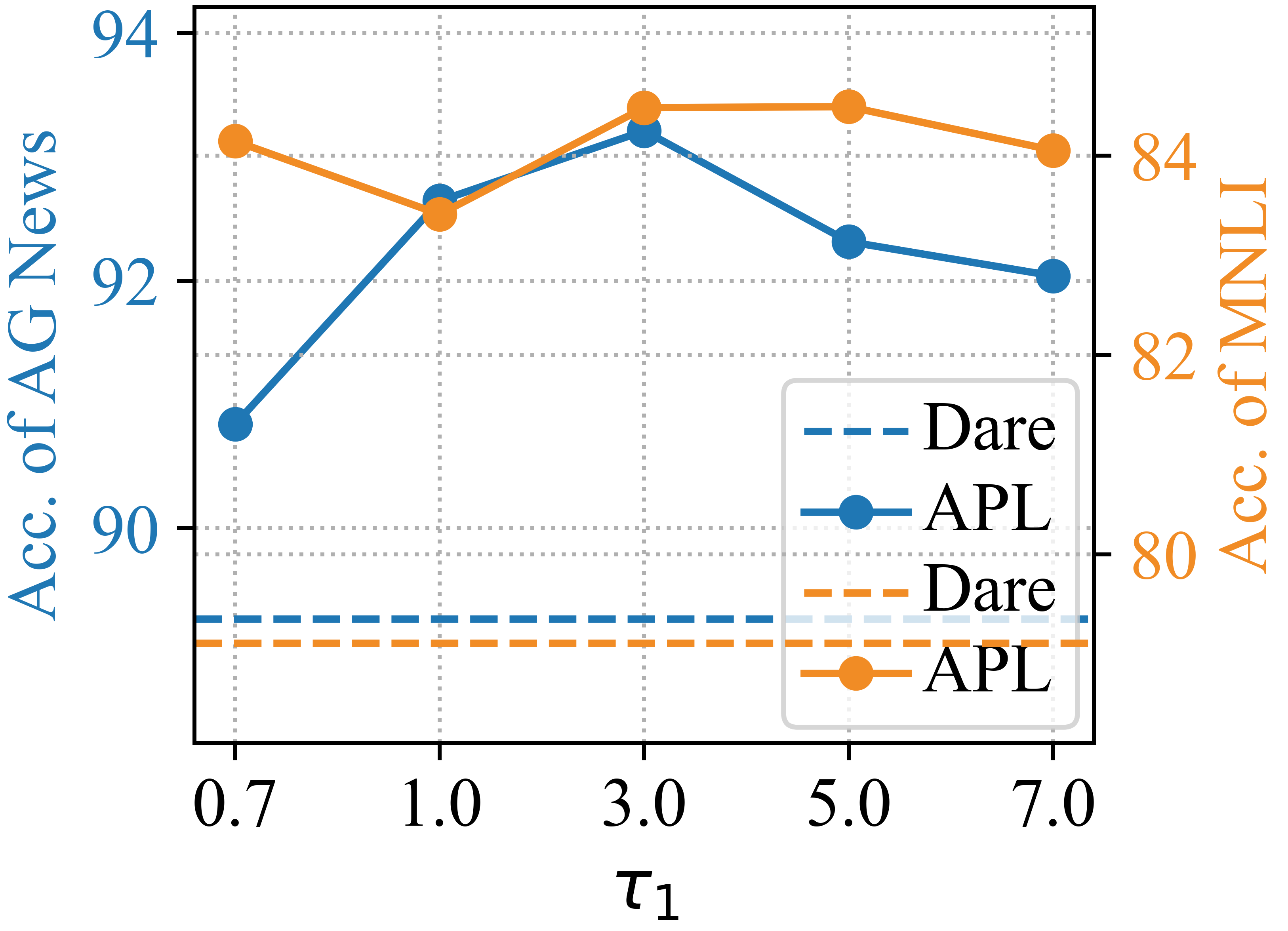} &  \hspace{-10pt}\includegraphics[width=0.47\columnwidth]{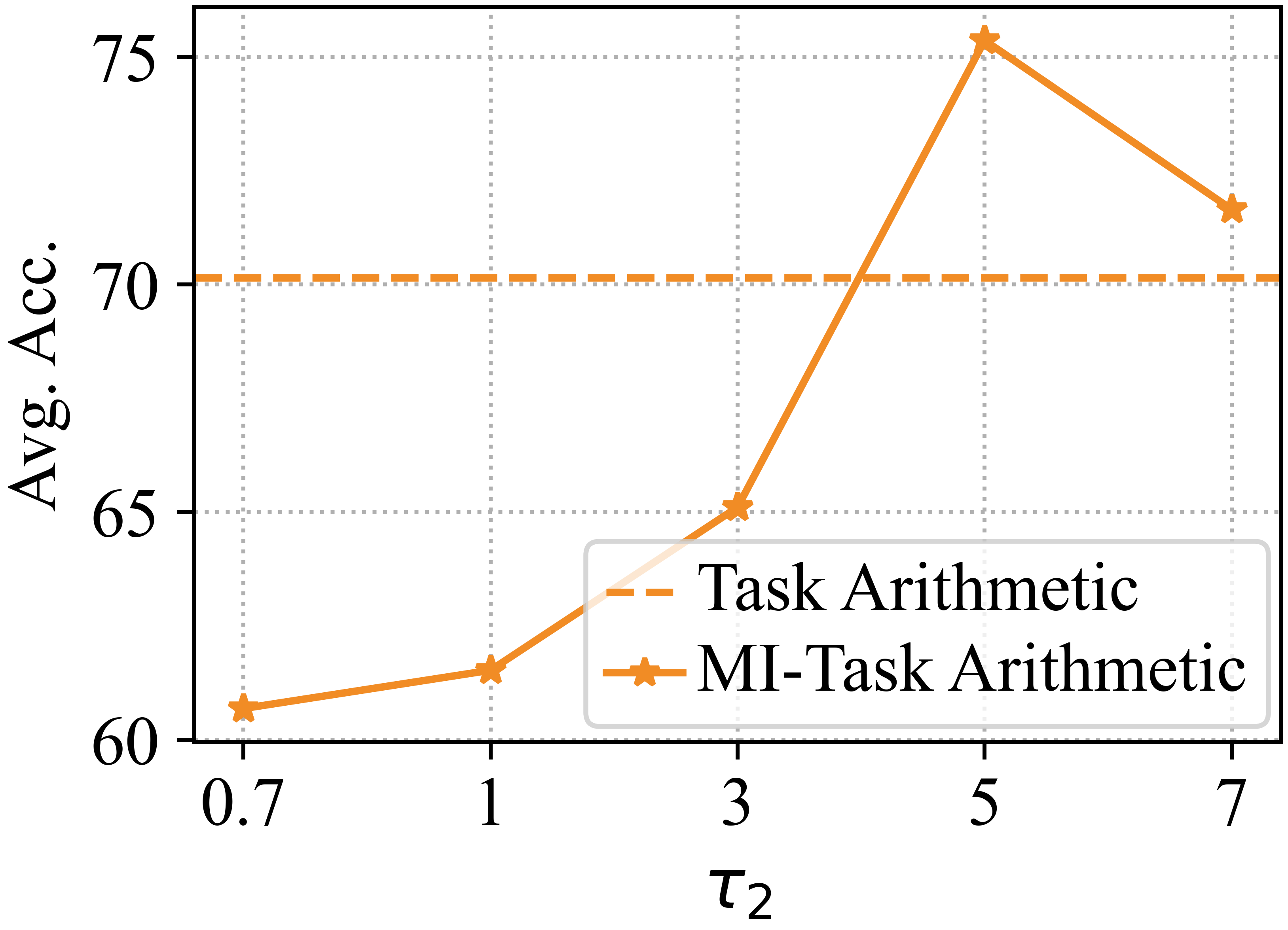} \\
    \end{tabular}
    \caption{Impact of different temperatures.}
    \label{fig:temperature}
\end{figure}
\begin{figure}[h]
    \centering
    \includegraphics[width=\linewidth]{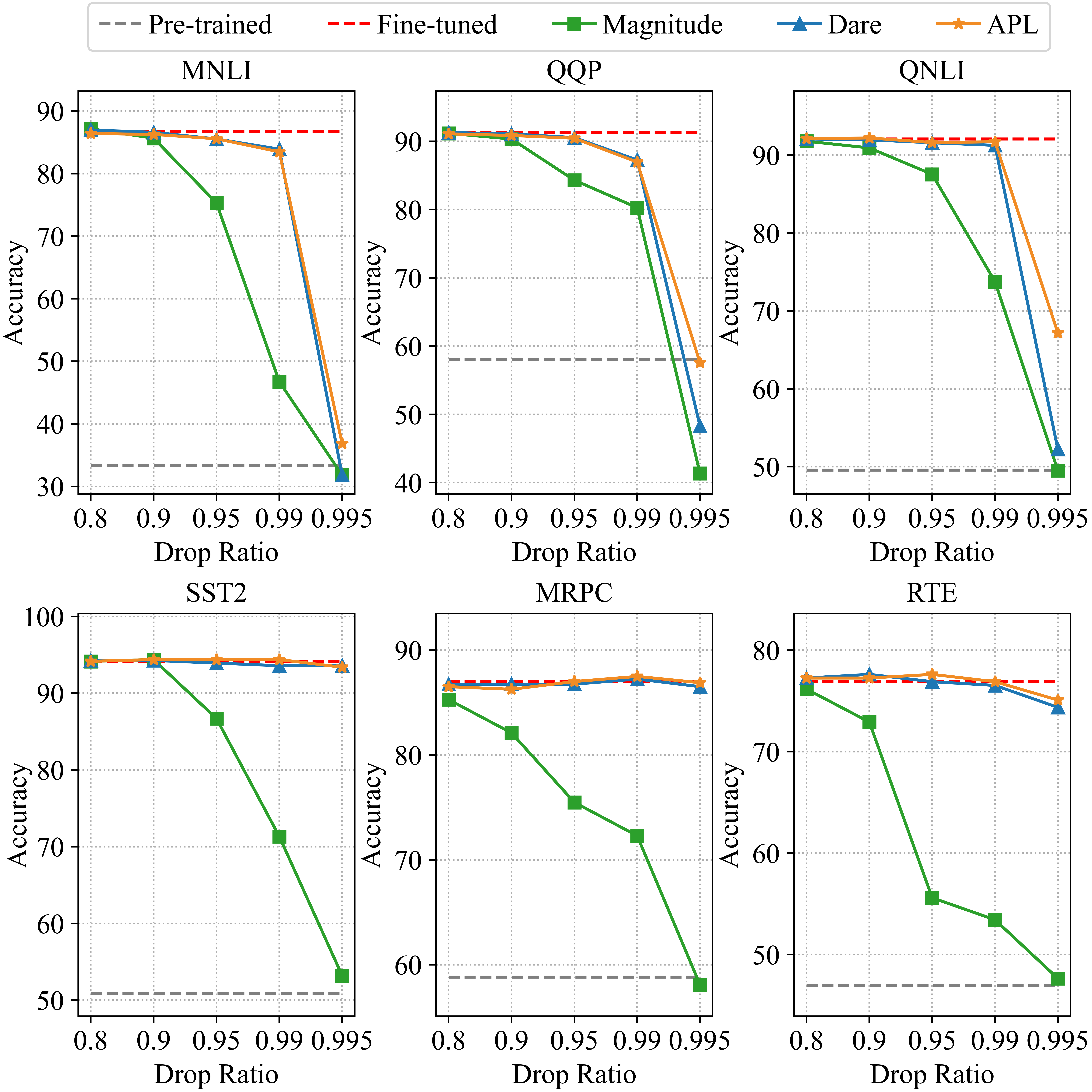}
    \caption{Pruning comparison on Roberta.}
    \label{fig:pruning-roberta}
\end{figure}

\subsection{Impact of Different Temperature}
\label{app:temperature}
Hyperparameters $\tau_1$ and $\tau_2$ control drop ratio and merging weight calibration, respectively. For $\tau_1$, we plot the accuracy of APL pruning on AG News and MNLI. For $\tau_2$, we plot the average accuracy (Avg. Acc.) in the in-domain merging setting. Results are shown in Figure \ref{fig:temperature}.

\subsection{Pruning Comparison on Roberta}
\label{app:roberta-pruning}
Except for the pruning comparison on the large language model Llama, we also conduct experiments on the encoder-based model Roberta-base to evaluate APL. Results are shown in Figure \ref{fig:pruning-roberta}. Under Roberta, APL also showcases its effectiveness in parameter pruning.

\begin{table}
    \centering
    \resizebox{0.8\linewidth}{!}{
    \begin{tabular}{c|cc|cc}
    \toprule
    \multirow{2}{*}{Dataset} & \multicolumn{2}{c|}{Layer-level} & \multicolumn{2}{c}{Hidden State-level}\\
     & APL & w/GA & APL & w/GA \\
    \midrule
    AGNews & 89.86 & 90.63 & 92.84 & 93.26 \\
    Hellaswag & 58.31 & 60.23 & 61.32 & 59.16 \\
    MNLI & 84.27 & 80.28 & 82.12 & 80.88 \\
    MRPC & 86.76 & 83.33 & 87.50 & 85.05 \\
    SST2 & 92.77 & 94.38 & 93.23 & 94.84 \\
    Winogrande & 68.19 & 70.63 & 68.75 & 69.85 \\
    \midrule
    Average & 80.03 & 79.91 & 80.96 & 80.51 \\
    \bottomrule
    \end{tabular}
    }
    \caption{Ablation study of gradient approximation.}
    \label{tab:gradient-approximation}
\end{table}
\subsection{Impact of Gradient Approximation}
To evaluate the influence of gradient approximation, we compare the standard APL (APL) and APL with gradient approximation (w/GA) in Table \ref{tab:gradient-approximation}. Generally, hidden state-level methods achieve better performance than layer-level, which implies fine-grained parameter partition brings enhancement in parameter pruning. Further, w/GA performs lower but considerably performance compared with APL, showcasing gradient approximation could be a promising substitute with APL for efficiency ensuring.

\subsection{Merging on Two Tasks}
\label{app:merge-two}
To compare each model under a simple setting where the number of merging models is two, we selected a similar task pair, AG News and SST2, and a dissimilar task pair, AG News and Winogrande, for comparison.
Results are shown in Table \ref{tab:in-domain-2-task}. We observe that under this simple setting, regardless of task similarity, each merging method can achieve results that are convincingly close to those of the Fine-tuned Model. Previous works have already proved this. In the context of merging six models, where parameter conflicts are more pronounced, we find that APL performs better in this harder scenario. This demonstrates the effectiveness of APL in reducing redundant parameters.
\begin{table}
    \centering
    \resizebox{\linewidth}{!}{
    \begin{tabular}{c|cc|cc}
        \toprule
        Model & AG News & SST2 & AG News & Winogrande \\
         \midrule
        Fine-tuned Model & 94.42 & 95.53 & 94.42 & 75.45 \\
        \midrule
        Task Arithmetic & 94.17 & 96.33 & 94.24 & 76.32 \\
        \textit{with Dare} & 94.36 & 96.21 & 94.28 & 75.85 \\
        \textit{with APL} & 94.33 & 96.10 & 94.17 & 76.48 \\
        \midrule
        MI-Task Arithmetic & 94.32 & 96.21 & 94.55 & 76.59 \\
        \textit{with Dare} & 94.32 & 96.44 & 94.55 & 76.48 \\
        \textit{with APL} & 94.37 & 96.10 & 94.47 & 76.80 \\
         \bottomrule
    \end{tabular}
    }    
    \caption{In-domain merging results on the two tasks' setting.}
    \label{tab:in-domain-2-task}
\end{table}

\end{document}